\documentclass[runningheads,a4]{llncs}
\usepackage{amssymb}
\usepackage{float}
\usepackage{graphicx}
\usepackage{subfigure}
\usepackage{algorithm}
\usepackage{algorithmicx}
\usepackage{algpseudocode}
\usepackage{amsmath}
\usepackage{CJK}
\usepackage{verbatim}
\usepackage{url}
\begin{document}
\title{Monte Carlo Q-learning for General Game Playing}
%
%
\author{Hui Wang, Michael Emmerich, Aske Plaat}
\authorrunning{Hui Wang et al.}
%
\institute{Leiden Institute of Advanced Computer Science, Leiden University,\\ Leiden, the Netherlands\\
\email{h.wang.13@liacs.leidenuniv.nl}\\
\url{http://www.cs.leiden.edu}}
\maketitle              
%
\begin{abstract}
  After the recent groundbreaking results of AlphaGo, we have seen a strong interest in reinforcement learning in game playing. General Game Playing (GGP) provides a good testbed for reinforcement learning. In GGP, a specification of games rules is given.
  GGP problems can be solved by reinforcement learning. Q-learning is one of the canonical reinforcement learning methods, and has been used by (Banerjee $\&$ Stone, IJCAI 2007) in GGP.
  In this paper we implement Q-learning in GGP for three small-board games (Tic-Tac-Toe, Connect Four, Hex), to allow comparison to Banerjee et al. As expected, Q-learning converges, although  much slower than MCTS. Borrowing an idea from MCTS,
  we enhance Q-learning with Monte Carlo Search, to give QM-learning.
  This enhancement improves the performance of pure Q-learning.
  We believe that QM-learning can also be used to improve performance of reinforcement learning further for larger games, something which we will test in future work.

\end{abstract}

\section{Introduction}
Traditional game playing programs are written to play a single specific game, such as Chess, or Go. The aim of {\em General\/} Game Playing~\cite{Genesereth2005} (GGP) is to create adaptive game playing programs; programs that can play more than one game well. To this end, GGP applies a so-called Game Description Language (GDL)~\cite{Love2008}.
GDL-authors write game-descriptions that specify the rules of a game. The challenge for GGP-authors is to write a GGP player that will play  any game well.
GGP players should ensure that a wide range of GDL-games can be run efficiently. Comprehensive tool-suites exist to help researchers write GGP and GDL programs, and an active research community exists~\cite{Kaiser2007,Genesereth2014,Swiechowski2014}.

The GGP model follows the state/action/result paradigm of \mbox{reinforcement} learning~\cite{Sutton1998}, a paradigm that has yielded many successful problem solving algorithms. For example, the recent successes of AlphaGo are based on two reinforcement learning algorithms, Monte Carlo Tree Search (MCTS)~\cite{Browne2012} and Deep Q-learning (DQN)~\cite{Mnih2015,Silver2016}. MCTS, in particular, has been successful in GGP~\cite{Mehat2008}.

The AlphaGo successes have also shown that  for Q-learning much compute power is needed, something already noted in Banerjee \cite{Banerjee2007}, who reported slow convergence for Q-learning. Following Banerjee, in this paper we address the convergence speed  of Q-learning. We use three 2-player zero-sum games: Tic-Tac-Toe, Hex and Connect Four, and table-based Q-learning. Borrowing an idea from MCTS, we then create a new version of Q-learning,\footnote{Despite the success of deep-learning techniques in the field of game playing, we consider it to be valuable to develop more light-weight, table-based, machine learning techniques for smaller board games. Such light-weight techniques would have the advantage of being more accessible to theoretical analysis and of being more efficient with respect to computational resources.} inserting Monte Carlo Search (MCS) into the Q-learning loop.


Our contributions can be summarized as follows:
\begin{enumerate}
\item We evaluate the classical Q-learning algorithm, finding (1) that Q-learning works in GGP, and (2) that classical Q-learning converges slowly in comparison to MCTS.
\item To improve performance, and in contrast to~\cite{Banerjee2007}, we enhance Q-learning by adding a modest amount of Monte Carlo lookahead (QMPlayer)~\cite{Robert2004}. This improves the rate of convergence of Q-learning.
\end{enumerate}

The paper is organized as follows. Section 2 presents  related work and recalls basic concepts of GGP and reinforcement learning. Section 3 provides the design of a player for single player games. We further discuss the player to play two-player zero-sum games, and implement such a player (QPlayer). Section 4 presents the player, inserting MCS, into Q-learning (QMPlayer). Section 5 concludes the paper and discusses directions for future work.

\section{Related Work and Preliminaries}
\subsection{GGP}
A General Game Player must be able to accept formal GDL descriptions of a game and play games effectively without human intervention~\cite{Genesereth2014}. A  Game Description Language (GDL) has been defined to describe the game rules ~\cite{Thielscher2011}. An interpreter program~\cite{Swiechowski2014} generates the legal moves (actions) for a specific board (state).
Furthermore, a Game Manager (GM) is at the center of the software ecosystem. The GM interacts with game players through the TCP/IP protocol to control the  match. The GM manages game descriptions and matches records and temporary states of matches while the game is running. The system also contains  a viewer interface for users who are interested in running matches and a monitor to analyze the match process.

\subsection{Reinforcement Learning}
Since Watkins proposed Q-learning in 1989~\cite{Watkins1989}, much progress has been made in reinforcement learning~\cite{Even-Dar2002,Hu2003}. However, only few works report on the use of Q-learning in GGP. In~\cite{Banerjee2007}, Banerjee and  Stone propose a method to create a general game player to study knowledge transfer, combining Q-learning and GGP. Their aim is to improve the performance of Q-learning by transferring the knowledge learned in one game to a new, but related, game. They found knowledge transfer with Q-learning to be expensive. In our work, instead, we use Monte Carlo lookahead to get knowledge directly, in a single game.

Recently, DeepMind published work on mastering Chess and Shogi by self-play with a deep generalized reinforcement learning algorithm \cite{Silver2017b}. With a series of landmark publications from AlphaGo to AlphaZero~\cite{Silver2016,Silver2017a,Silver2017b}, these \mbox{works} showcase the promise of general reinforcement learning algorithms. However, such learning algorithms are very resource intensive and typically require special GPU hardware. Further more, the neural network-based approach is quite inaccessible to theoretical analysis. Therefore, in  this paper we study performance of table-based Q-learning.

In General Game Playing, variants of MCTS~\cite{Browne2012} are used with great success~\cite{Mehat2008}. M\'ehat et al. combined UCT and nested MCS for single-player general game playing~\cite{Mehat2010}. Cazenave et al. further proposed a nested MCS for two-player games~\cite{Cazenave2016}. Monte Carlo techniques have proved a viable approach for searching intractable game spaces and other optimization problems~\cite{Ruijl2014}. Therefore, in this paper we combine MCS to improve performance.

\subsection{Q-learning}
A basic distinction of reinforcement learning methods is that of  "on-policy" and "off-policy" methods. On-policy methods attempt to evaluate or improve the policy that is used to make decisions, whereas off-policy methods evaluate or improve a policy {\em different\/} from that used to make decisions~\cite{Sutton1998}. Q-learning is an off-policy method. The reinforcement learning model consists of an \emph{agent}, a set of states $S$, and a set of actions $A$ of every state $s$, $s\in S$~\cite{Sutton1998}. In Q-learning, the agent can move to the next state $s^\prime$, $s^\prime\in S$ from state $s$ after following action $a$, $a\in A$, denoted as $s\xrightarrow{a}s^\prime$. After finishing the action $a$, the agent gets a reward, usually a numerical score, which is to be maximized (highest reward). In order to achieve this goal, the agent must find the optimal action for each state. The reward of current state $s$ by taking the action $a$, denoted as $Q(s, a)$, is a weighted sum, calculated by the immediate reward $R(s, a)$ of moving to the next state and the maximum expected reward of all future states' rewards:
\begin{equation}
Q(s,a)=R(s,a)+\gamma (max_{a^\prime}Q(s^\prime,a^\prime))
\end{equation}
where $a^\prime \in A^\prime$, $A^\prime$ is the set of actions under state $s^\prime$. $\gamma$ is the discount factor of $max Q(s^\prime,a^\prime)$ for next state $s^\prime$. $Q(s,a)$ can be updated by online interactions with the environment using
the following rule:
\begin{equation}
Q(s,a)\leftarrow (1-\alpha)Q(s,a)+\alpha( R(s,a)+\gamma (max_{a^\prime}Q(s^\prime,a^\prime)))
\end{equation}
where $\alpha \in [0,1]$ is the learning rate. The Q-values are guaranteed to converge by some schemas, such as exploring every $(s, a)$, which should be ensured by a suitable exploration and exploitation method (such as $\epsilon$-greedy).

\section{Classical Q-learning}
\subsection{Exploration/Exploitation: $\epsilon$-greedy}
\begin{figure}[H]
\centering
\includegraphics[width=0.75\textwidth]{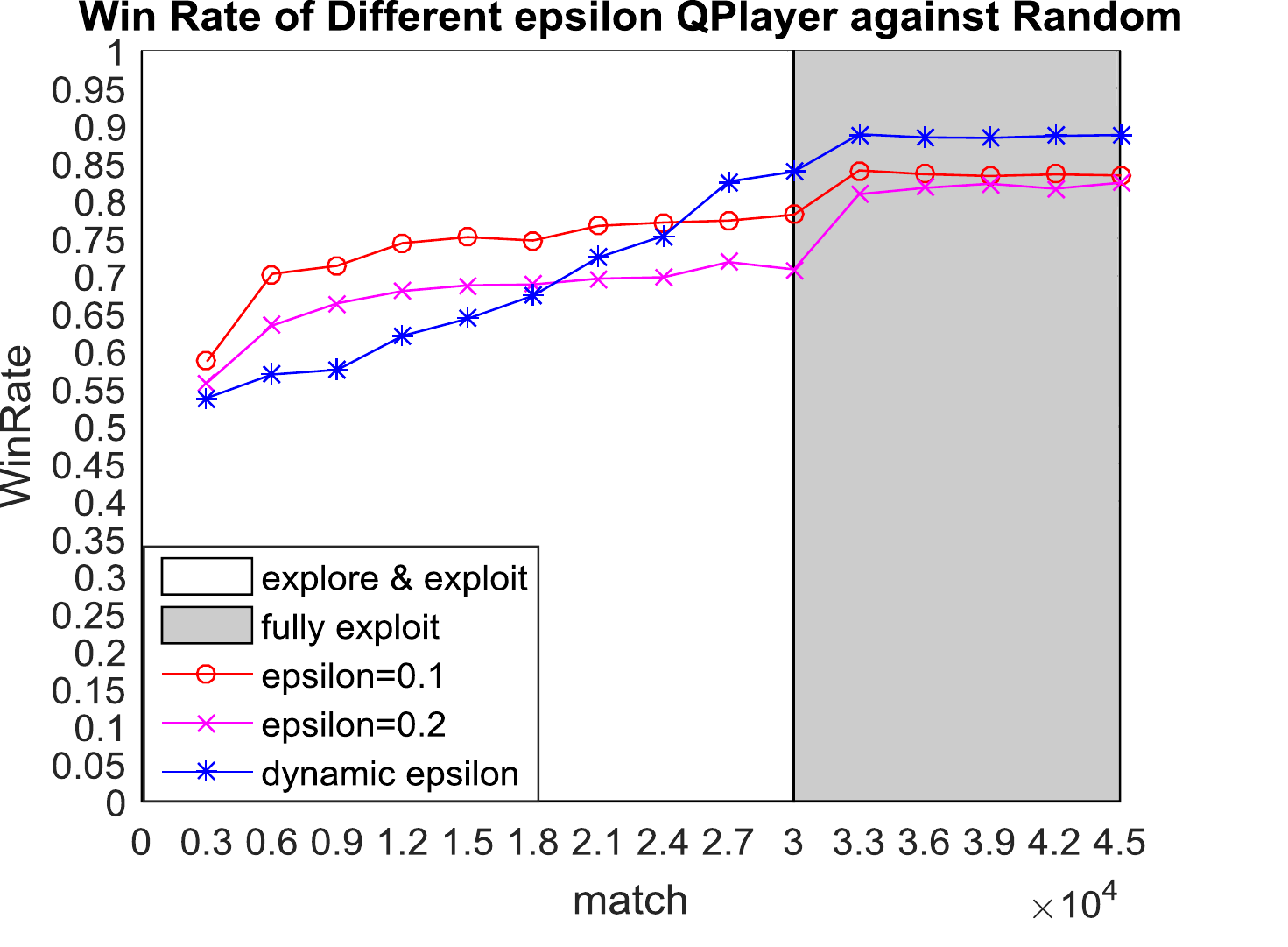}
\caption{Win Rate of Fixed and Dynamic $\epsilon$ Player vs Random in Tic-Tac-Toe. In the white part, the player uses $\epsilon$-greedy to learn; in the grey part, all players set $\epsilon$=0 (stable performance)}
\label{fig:figfixedepsilon} 
\end{figure}

As our baseline we use $\epsilon$-greedy Q-learning~\cite{Even-Dar2002} to balance exploration and exploitation. In order to find a better baseline player, we create $\epsilon$-greedy Q-learning players($\alpha=0.1$, $\gamma=0.9$) with fixed $\epsilon$=0.1, 0.2 and dynamically decreasing $\epsilon \in [0, 0.5]$ to play 30000 matches against Random player, respectively. During these 30000 matches, dynamic $\epsilon$  decreases from 0.5 to 0, fixed $\epsilon$ are 0.1, 0.2, respectively. After 30000 matches, fixed $\epsilon$ is also set to 0 to continue the competition. Results in Fig.\ref{fig:figfixedepsilon} show that dynamically decreasing $\epsilon$  performs better. We see that the final win rate of dynamically decreasing $\epsilon$ is 4\%  higher than fixed $\epsilon$=0.1 and 7\% higher than fixed $\epsilon$=0.2.

\begin{figure}[H]
\centering
\includegraphics[width=0.75\textwidth]{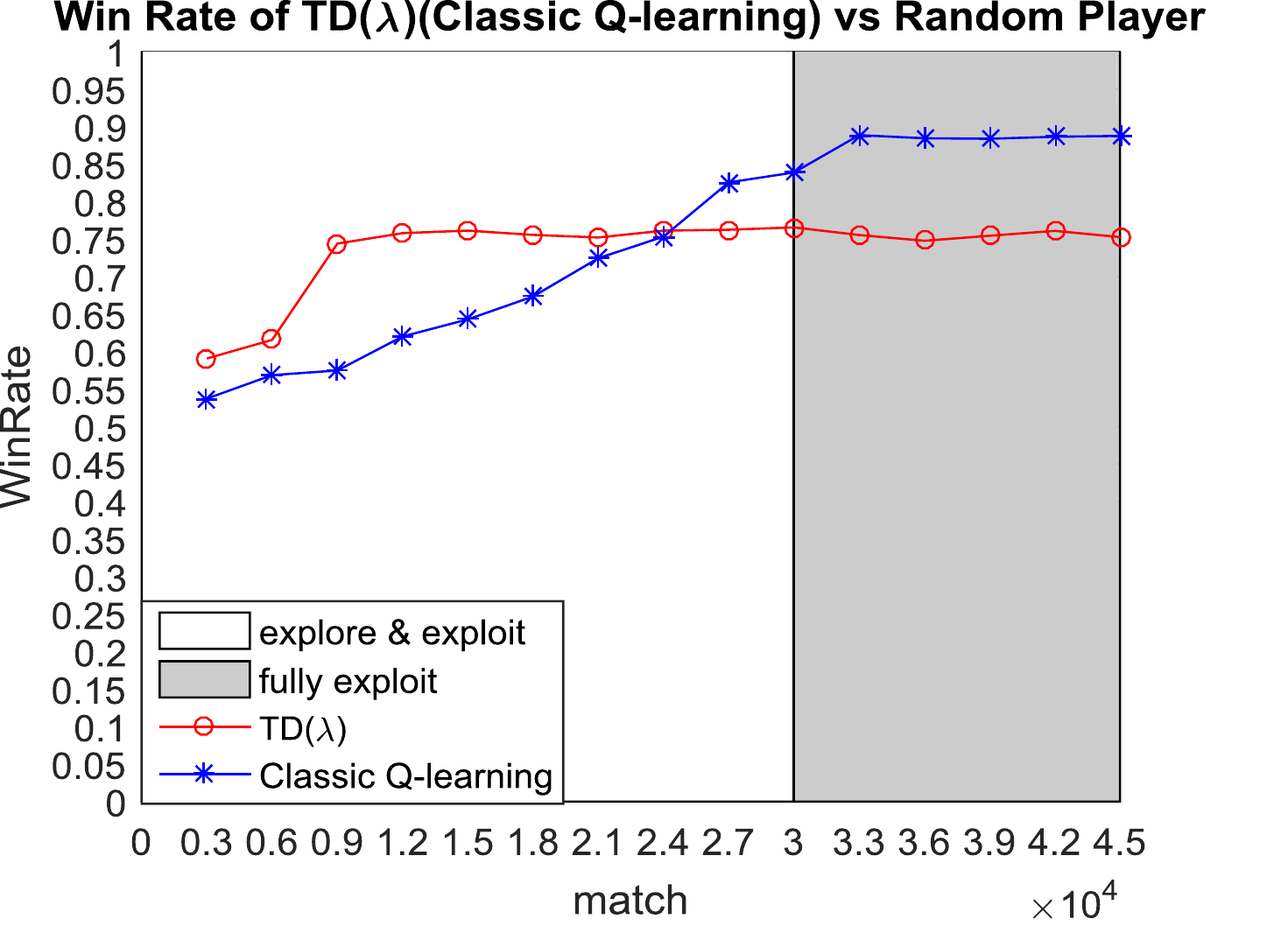}
\caption{Win Rate of $\epsilon$-greedy Q-learning and [11] Baseline Player vs Random in Tic-Tac-Toe. In the white part, the player uses $\epsilon$-greedy to learn; in the grey part, all players set $\epsilon$=0 (stable performance)}
\label{fig:figbaseline} 
\end{figure}

To enable comparison with previous work, we compare TD($\lambda$), the baseline learner of~\cite{Banerjee2007}($\alpha=0.3$, $\gamma=1.0$, $\lambda=0.7$, $\epsilon=0.01$), and our baseline learner($\alpha=0.1$, $\gamma=0.9$, $\epsilon \in [0, 0.5]$, Algorithm 1). For Tic-Tac-Toe, from Fig.\ref{fig:figbaseline}, we find that although the TD($\lambda$) player converges more quickly initially (win rate stays at about 75.5$\%$ after 9000th match), our baseline performs better when the value of $\epsilon$-greedy decreases dynamically with  the learning process.

In our dynamic implementation, we use the function
$$\epsilon(m)=\begin{cases}
a(\cos(\frac{m}{2l}\pi))+b & m\leq l\\
0& m>l
\end{cases}$$
for $\epsilon$, where $m$ is the current match count, $l$ is the total learning match, which we set in advance. $a$ and $b$ can be set  to limit the range of $\epsilon$, where $\epsilon\in[b,a + b]$, $a, b \geq 0$ and $a + b \leq1$.
The player generates a random number $num$ where $num \in[0,1]$. If $num < \epsilon$, the player will explore a new action randomly, else the player will choose a best action from the currently learnt $Q(s,a)$ table.

The question is what to do when the player cannot get a record from $Q(s,a)$? In classical Q-learning, it chooses an action randomly (Algorithm~\ref{alg:algorithm4}). While in the enhanced algorithm, we insert MCS (Algorithm 3) inside Q-learning, giving  Algorithm~\ref{alg:algorithm6}.  QMPlayer combines MCS and Random strategies in the first part of the search, but after enough state-action pairs are learned, it performs just like QPlayer.


\subsection{Q-learning  for Single Player Games}
We start by introducing the simplest algorithm, playing single player games. Since games are played by only one player, we just need to build one $Q(s,a)$ table for the player to select the best action under the specific state, see Algorithm 1~\cite{Sutton1998}:
\allowdisplaybreaks
\begin{algorithm}[H]
\renewcommand{\algorithmicrequire}{ \textbf{Input:}} 
\renewcommand{\algorithmicensure}{ \textbf{Output:}} 
\caption{Basic Q-learning Player For Single Player Games}
\label{alg:algorithm2}
\begin{algorithmic}[1] 
\Require ~~\\ 
game state: \emph{S};\\
legal actions:\emph{A}\\
learning rate: $\alpha$\\
discount factor: $\gamma$;\\
reward: $R(S,A)$;\\
updating table: $Q(S,A)$;
\Ensure ~~\\ 
selected action according to updating table: $Q(S,A)$;
\Function{BasicQlearningSingle}{$S,A$}
\For{each learning match}
\For{each game state during match}
\State Update $Q(s,a)\leftarrow (1-\alpha)Q(s,a)+\alpha(R(s,a)+\gamma max_{a^\prime}Q(s^\prime , a^\prime))$;
\EndFor
\EndFor
\State selected\ =\ false;
\State expected\_score\ =\ 0;
\For{each $q(s,a)$ in $Q(S,A)$}
\State if(current game state equals s and expected\_score\ $<$\ q(s,a));
\State expected\_score\ =\ q(s,a);
\State selected\_action\ =\ a;
\State selected\ =\ true;
\EndFor
\If{selected\ ==\ false}
\State selected\_action\ =\ Random();
\EndIf
\State \Return selected\_action;  
\EndFunction
\end{algorithmic}
\end{algorithm}

\subsection{Q-learning for Two-Player Games}
Next, we consider more complex games played by two players. In GGP, the  \emph{switch-turn} command allows every player to play the game roles by turn. 
The Q-learning player should build corresponding $Q(s,a)$ tables for each role.

Since our GGP games are  two-player zero-sum games, we can use the same rule, see Algorithm~\ref{alg:algorithm4} line~\ref{alg:algorithm4:rule}, to create $R(s,a)$ rather than to use a reward table.
In our experiments, we set $R(s,a)=0$ for non-terminal states, and call the $getGoal()$ function for terminal states. 
In order to improve the learning effectiveness, we update the corresponding $Q(s,a)$ table only at the end of the match.
\allowdisplaybreaks
\begin{algorithm}[H]
\renewcommand{\algorithmicrequire}{ \textbf{Input:}} 
\renewcommand{\algorithmicensure}{ \textbf{Output:}} 
\caption{$\epsilon$-greedy Q-learning Player For Two-Player Zero-Sum Games}
\label{alg:algorithm4}
\begin{algorithmic}[1] 
\Require ~~\\ 
game state: \emph{S};\\
legal actions:\emph{A};\\
learning rate: $\alpha$;\\
discount factor: $\gamma$;\\
corresponding updating tables: $Q_{myrole}(S,A)$ for every role in the game;
\Ensure ~~\\ 
selected action according to updating table: $Q_{myrole}(S,A)$;
\Function{epsilonGreedyQlearning}{$S,A$}
\If{$\epsilon$-greedy is enabled}
\For{each learning match}
\State record\ =\ getMatchRecord();
\For{each state from termination to the beginning in record}
\State myrole\ =\ getCurrentRole();
\State R(s,a)\ =\ getReward(s,a);//$s^\prime$ is terminal state?\ getGoal($s^\prime$,myrole):0\label{alg:algorithm4:rule}
\State Update $Q_{myrole}(s,a)\leftarrow (1-\alpha)Q_{myrole}(s,a)+\alpha(R(s,a)+\gamma max_{a^\prime}Q_{myrole}(s^\prime , a^\prime))$;
\EndFor
\EndFor
\State selected\ =\ false;
\State expected\_ score\ =\ 0;
\For{each $q_{myrole}(s,a)$ in $Q_{myrole}(S,A)$}
\State if(current game state equals s and expected\_ score$\ <q_{myrole}(s,a))$;
\State expected\_ score\ =\ $q_{myrole}(s,a)$;
\State selected\_ action\ =\ a;
\State selected\ =\ true;
\EndFor
\If{selected\ ==\ false}
\State \emph{\textbf{selectedaction\ =\ Random()}};  \label{alg:algorithm4:random}
\EndIf
\Else
\State selected\_ action\ =\ Random()
\EndIf
\State \Return selected\_ action; 
\EndFunction
\end{algorithmic}
\end{algorithm}

\subsubsection*{Experiment 1} 
\begin{figure}[H]
\centering
\subfigure[total learning match=5000]{\label{fig:subfig2:a} 
\includegraphics[width=0.48\textwidth]{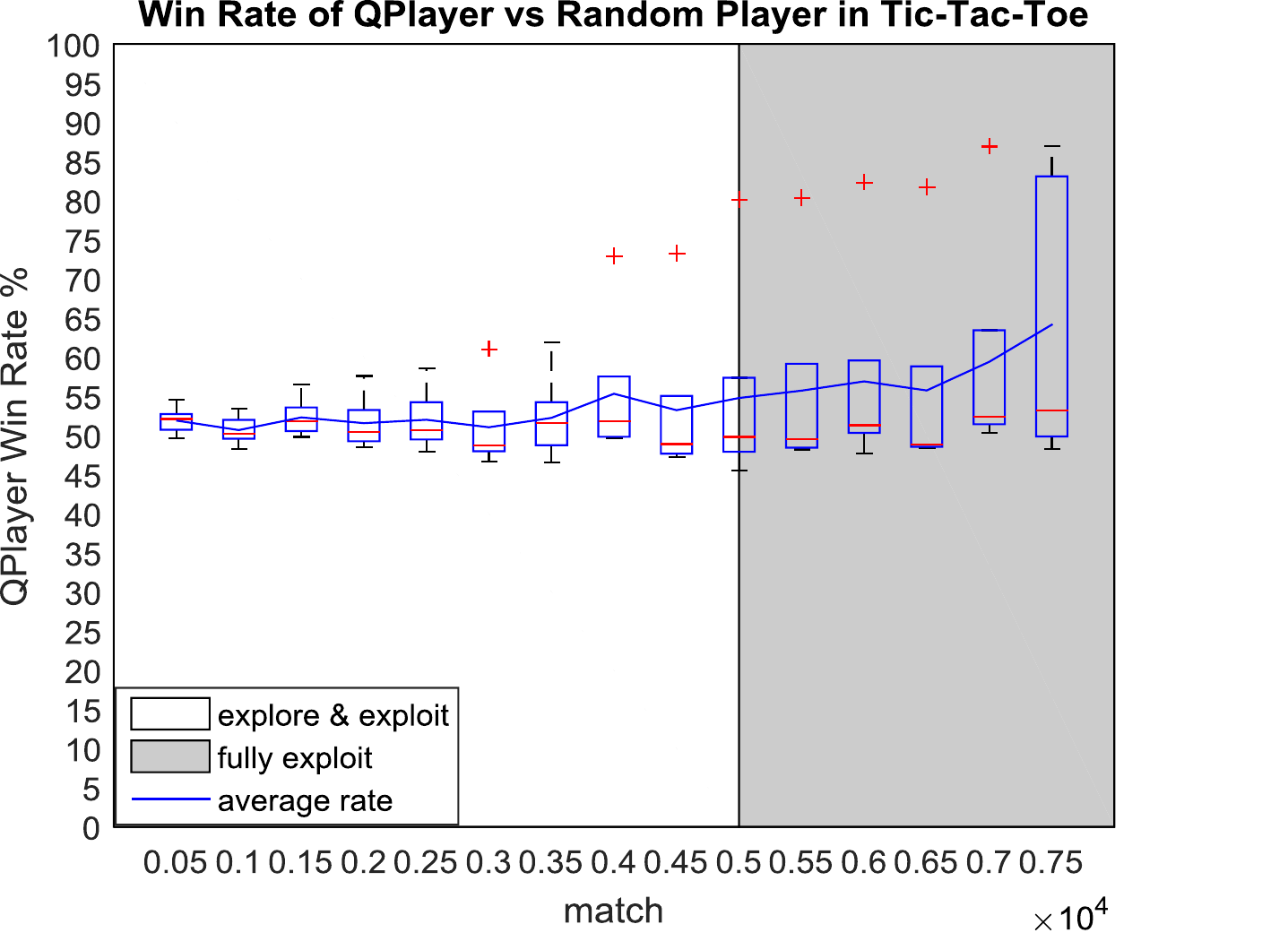}}
\hspace{0.000000000000001\textwidth}
\subfigure[total learning match=10000]{\label{fig:subfig2:b} 
\includegraphics[width=0.48\textwidth]{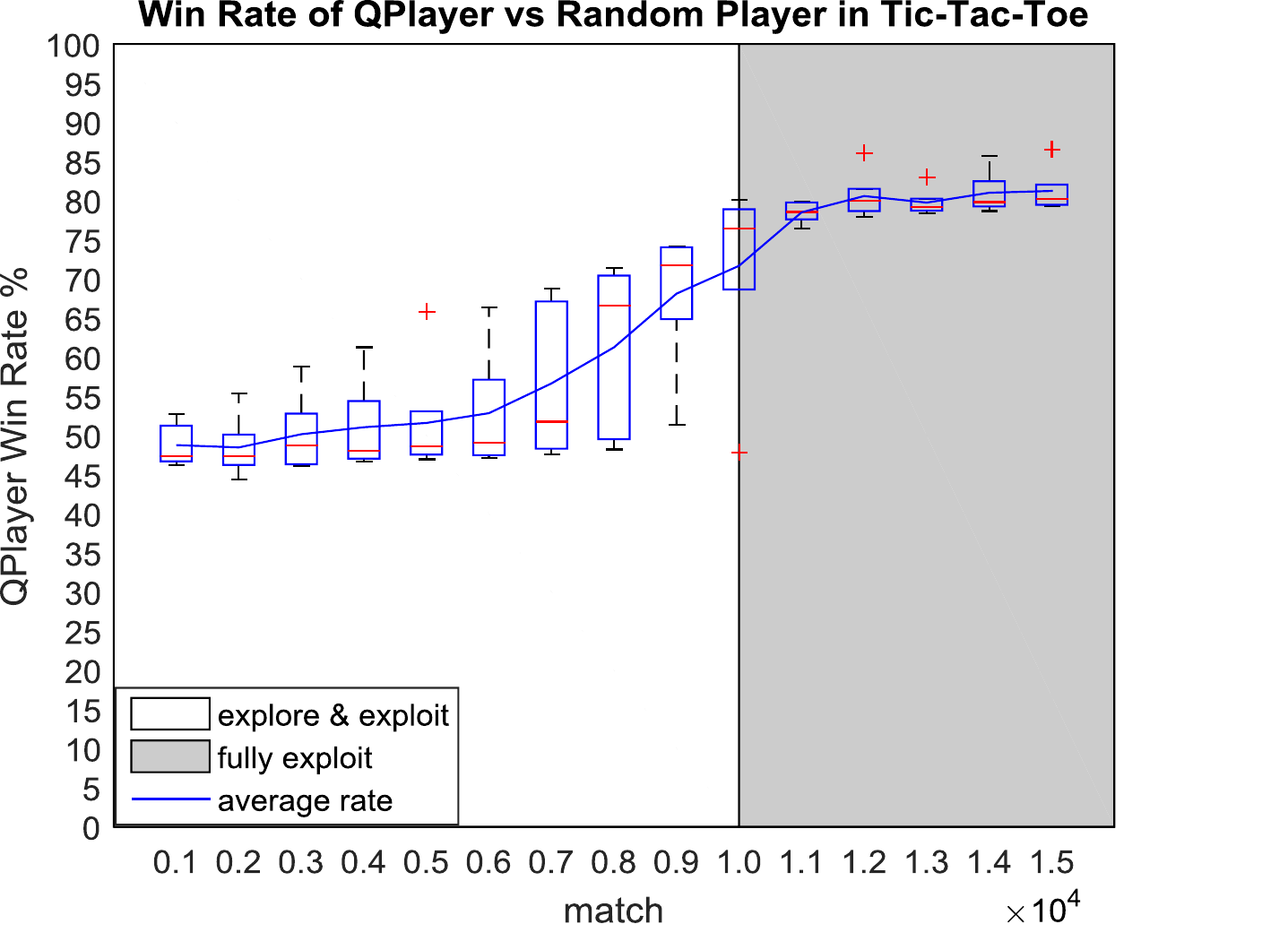}}
\hspace{0.000000000000001\textwidth}
\subfigure[total learning match=20000]{\label{fig:subfig2:c} 
\includegraphics[width=0.48\textwidth]{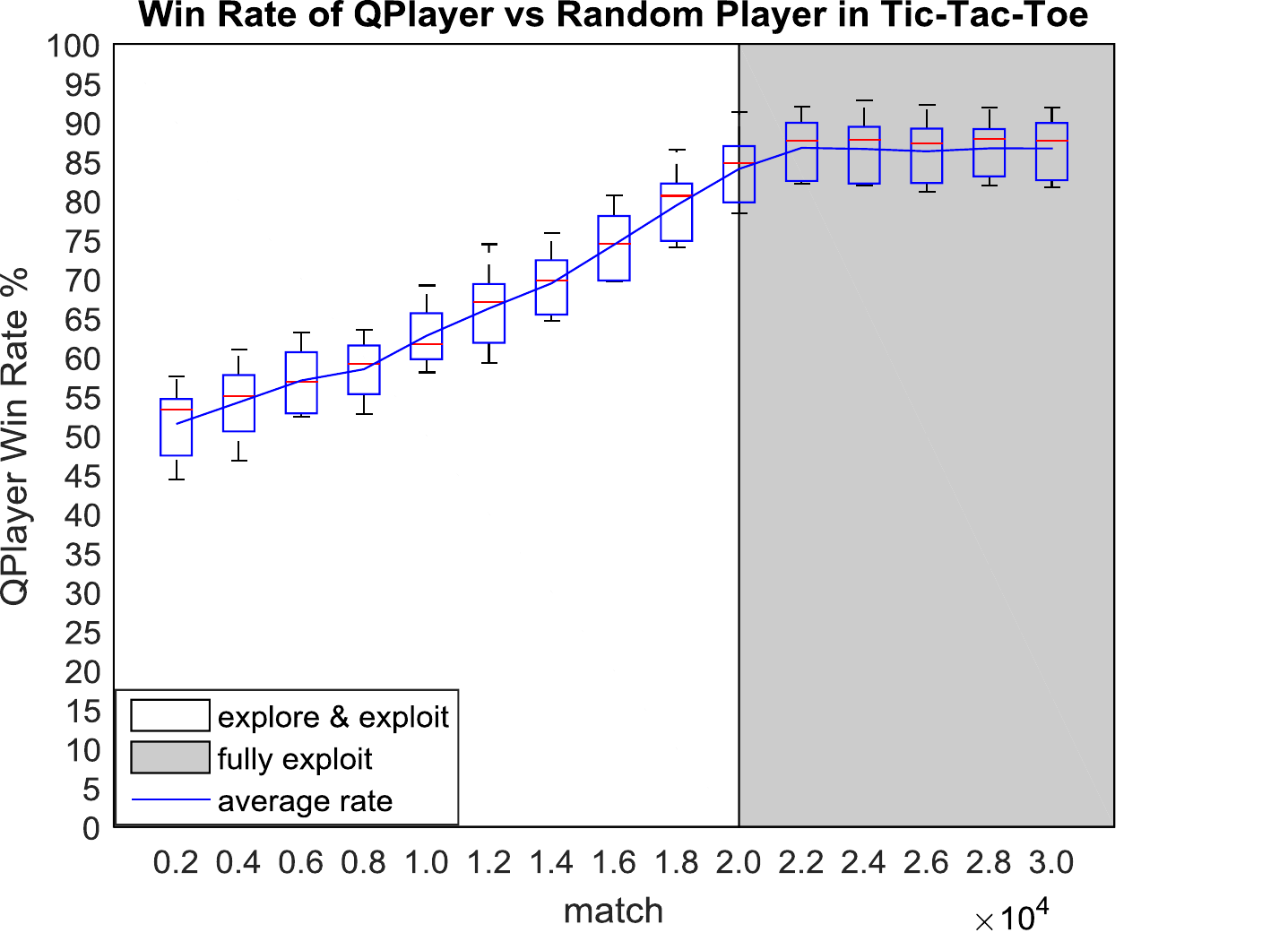}}
\hspace{0.000000000000001\textwidth}
\subfigure[total learning match=30000]{\label{fig:subfig2:d} 
\includegraphics[width=0.48\textwidth]{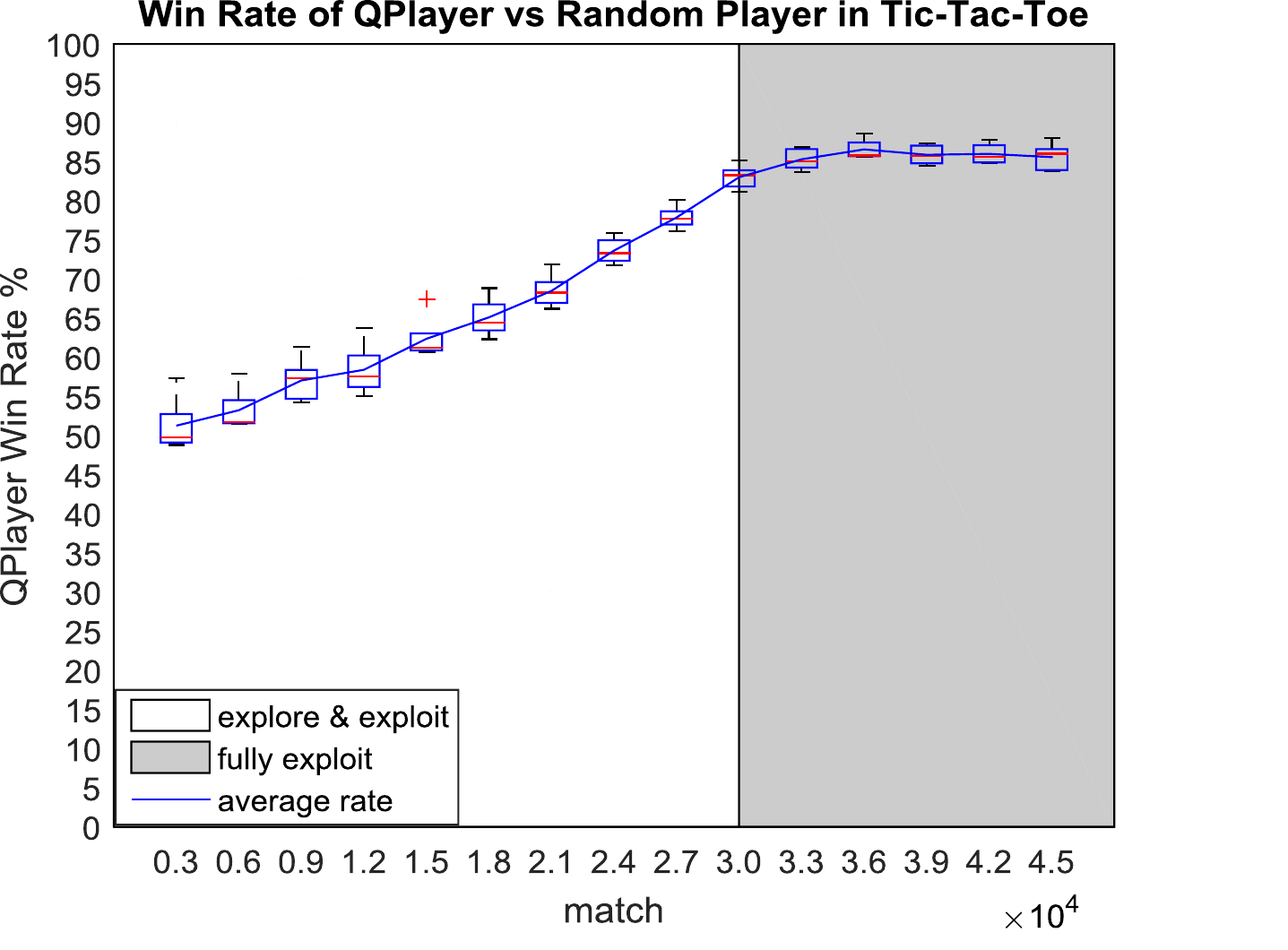}}
\hspace{0.000000000000001\textwidth}
\subfigure[total learning match=40000]{\label{fig:subfig2:e} 
\includegraphics[width=0.48\textwidth]{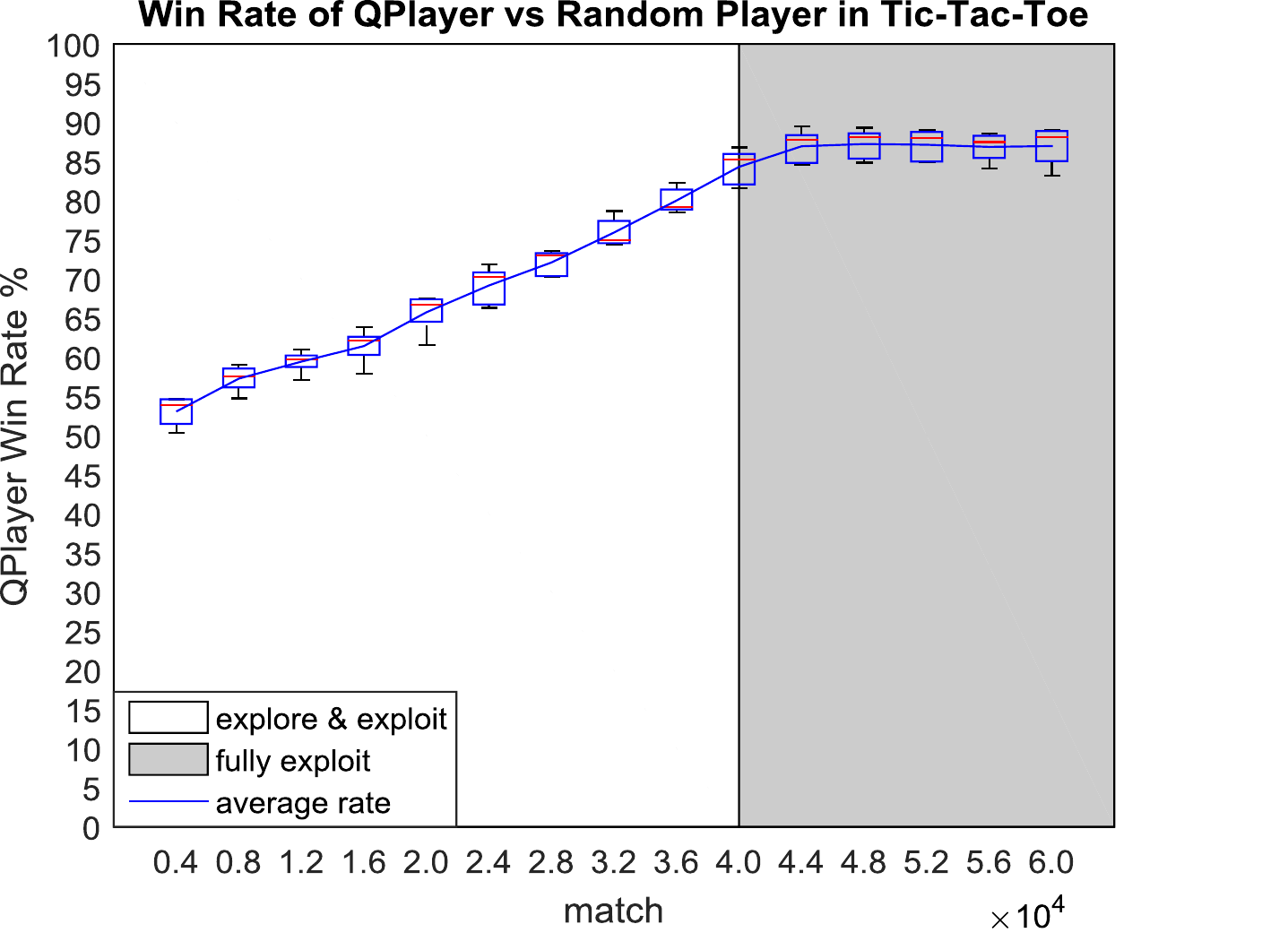}}
\hspace{0.000000000000001\textwidth}
\subfigure[total learning match=50000]{\label{fig:subfig2:f} 
\includegraphics[width=0.48\textwidth]{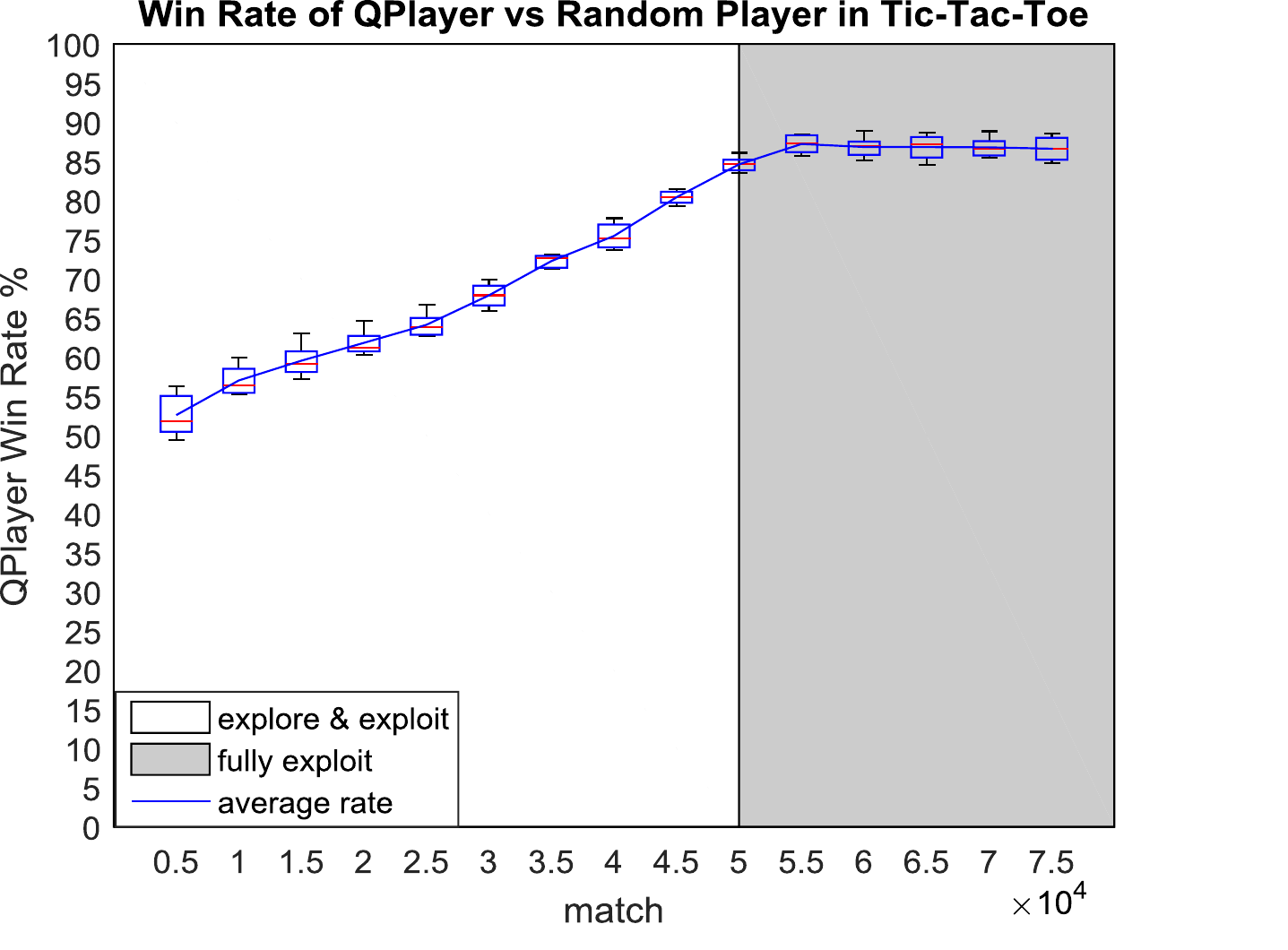}}
\caption{Win Rate of QPlayer vs Random Player in Tic-Tac-Toe averaged over 5 experiments. The winrate converges, and the variance is reduced as \emph{total learning match} increases}
\label{fig:subfig2} 
\end{figure}

In our first experiment we create QPlayer
(see Algorithm~\ref{alg:algorithm4}) to play Tic-Tac-Toe. We set parameters $\alpha=0.1$, $\gamma=0.9$, $\epsilon \in[0, 0.5]$,  respectively. As it learns to play Tic-Tac-Toe, we vary the {\em total learning match} in order to find how many matches it needs to learn to get convergence, and then make it play the game with Random player for $1.5\ \times\ ${\em total learning match} matches. We report results averaged over 5 experiments. The results are shown in Fig.\ref{fig:subfig2}.

%
Fig.\ref{fig:subfig2:a} shows  that QPlayer has the most unstable performance (the largest variance in 5 experiments)  and only wins around 55\% matches after training 5000 matches (i.e., 2500 matches trained for each role). Fig.\ref{fig:subfig2:b} illustrates that after training 10000 matches QPlayer wins about 80\% matches. However, during the exploration period (the light part of the figure) the performance is still very unstable. Fig.\ref{fig:subfig2:c} shows that QPlayer wins about 86\% of the matches while learning 20000 matches still with high variance. Fig.\ref{fig:subfig2:d}, Fig.\ref{fig:subfig2:e}, Fig.\ref{fig:subfig2:f}, show us that after training 30000, 40000, 50000 matches, QPlayer gets a similar win rate, which is nearly 86.5\% with smaller and smaller variance.

\begin{figure}[H]
\centering
\includegraphics[width=0.75\textwidth]{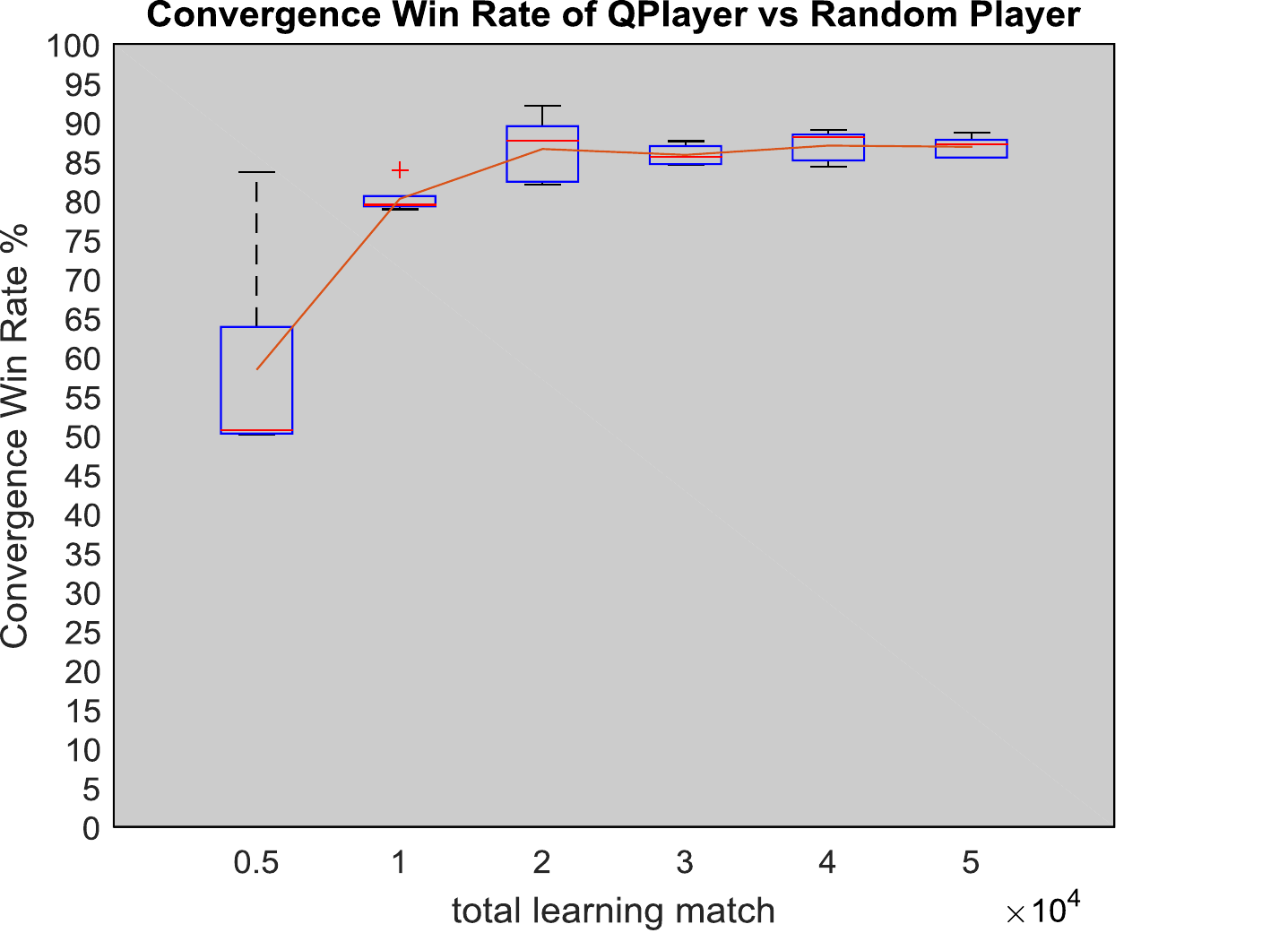}
\caption{Convergence Win Rate of QPlayer vs Random in Tic-Tac-Toe. Win rate converges as \emph{total learning match increases}}
\label{fig:figQfinalrate} 
\end{figure}

Overall, as the \emph{total learning match} increases, the win rate of QPlayer becomes higher until leveling off around 86.5\%. The variance becomes smaller and smaller. More intuitively, the QPlayer performance during the full exploitation period (the convergence results in the dark part of Fig.~\ref{fig:subfig2}) against different \emph{total learning match} is shown in Fig.\ref{fig:figQfinalrate}.

\begin{figure}[H]
\centering
\includegraphics[width=0.75\textwidth]{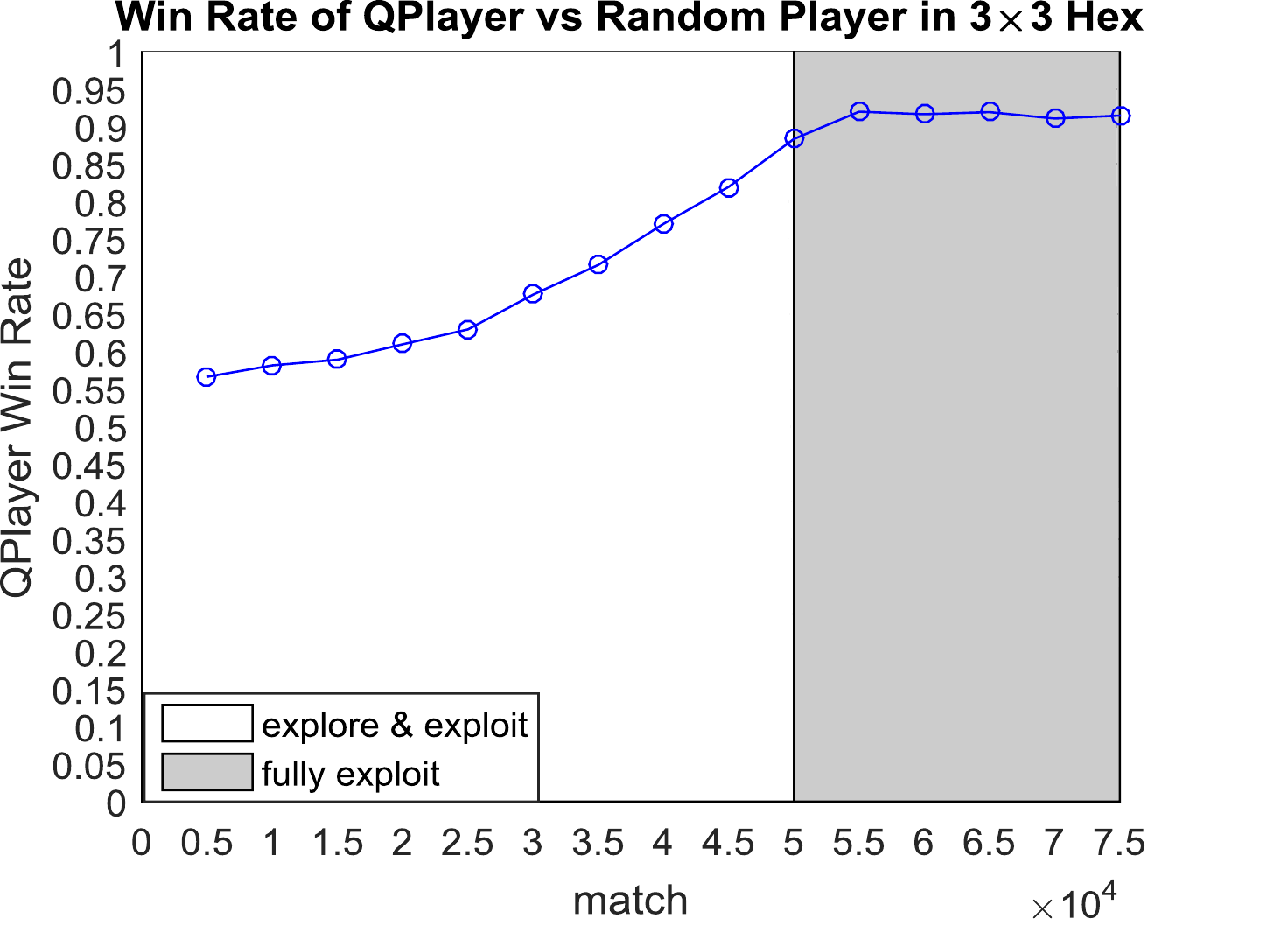}
\caption{Win Rate of QPlayer vs Random Player in 3$\times$3 Hex, the win rate of Q-learning also converges}
\label{fig:figHex} 
\end{figure}

\begin{figure}[H]
\centering
\includegraphics[width=0.75\textwidth]{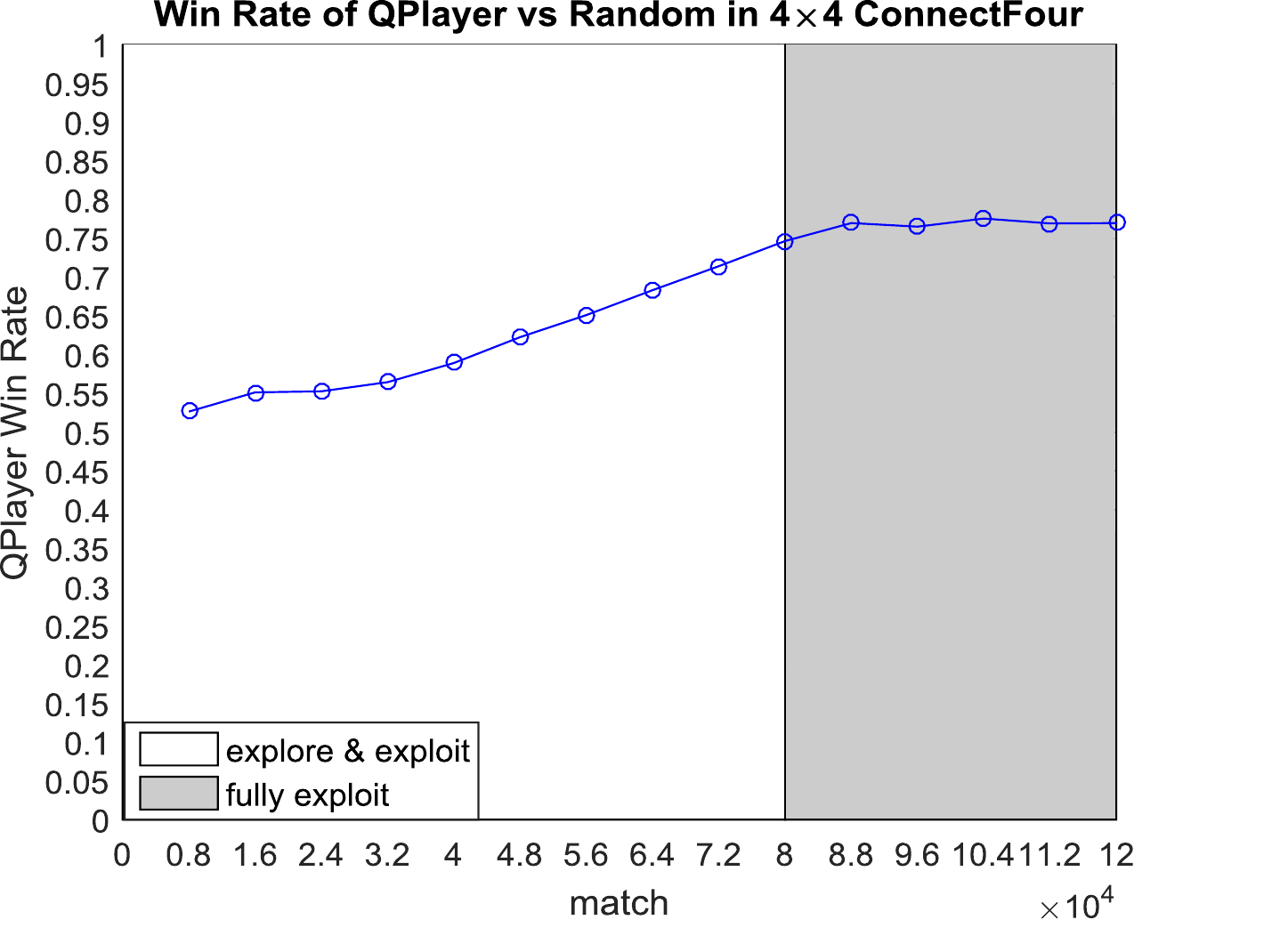}
\caption{Win Rate of QPlayer vs Random Player in 4$\times$4 ConnectFour, the win rate of Q-learning also converges}
\label{fig:figConnectFour} 
\end{figure}

Fig.\ref{fig:figQfinalrate} shows that QPlayer achieves, after convergence, a win rate of around 86.5\% with small variance. These experiments suggest indeed that Q-learning is applicable to a GGP system. However, beyond the basic applicability in a single game, we need to show that it can do so (1) {\em efficiently}, and (2) in more than one game. Thus, we further experiment with QPlayer to play Hex (learn 50000 matches) and Connect Four (learn 80000 matches) against the Random player. The results of these experiments are given in Fig.\ref{fig:figHex} and Fig.\ref{fig:figConnectFour}.

In order to limit excessive learning times, following~\cite{Banerjee2007}, we play Hex on a very small 3$\times$3 board, and play ConnectFour on a 4$\times$4 board. Fig.\ref{fig:figHex} and  Fig.\ref{fig:figConnectFour} show that QPlayer can also play these other games effectively.

However, there remains the problem that  QPlayer should be able to learn to play larger games. The complexity influences how many matches the QPlayer should learn. We show results to demonstrate how QPlayer performs while playing more complex games. We make QPlayer learn Tic-Tac-Toe 50000 matches (75000 for whole competition) in 3$\times$3, 4$\times$4, 5$\times$5 boards respectively and show the results in Fig.\ref{fig:figdifferentsize}:
\begin{figure}[H]
\centering
\includegraphics[width=0.75\textwidth]{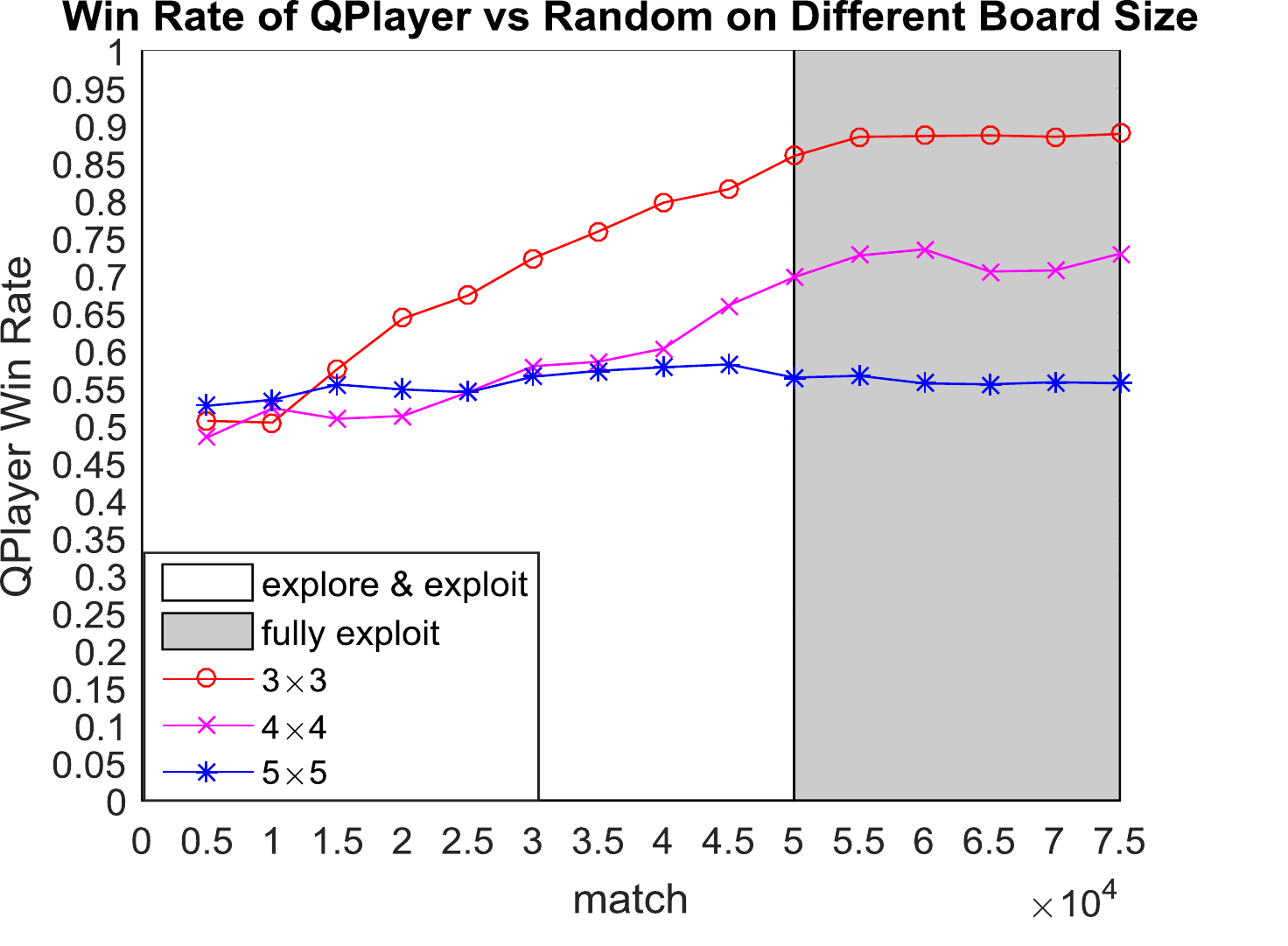}
\caption{Win Rate of QPlayer vs Random in Tic-Tac-Toe on Different Board Size. For larger board sizes convergence slows down}
\label{fig:figdifferentsize} 
\end{figure}
The results show that with the increase of game board size, QPlayer performs worse and for larger boards does not achieve convergence.

\section{Monte Carlo Q-learning}
\subsection{Monte Carlo Search}
The main idea of MCS~\cite{Robert2004} is to make some lookahead probes from a non-terminal state to the end of the game by selecting random moves for the players to estimate the value of that state. The pseudo code of time limited MCS in GGP is shown in Algorithm 3.
\allowdisplaybreaks
\begin{algorithm}[H]
\renewcommand{\algorithmicrequire}{ \textbf{Input:}} 
\renewcommand{\algorithmicensure}{ \textbf{Output:}} 
\caption{Time Limited Monte Carlo Search Algorithm}
\label{alg:algorithm1}
\begin{algorithmic}[1] 
\Require ~~\\ 
game state: \emph{S};\\
legal actions:\emph{A};\\
time limit of each searching $t$;
\Ensure ~~\\ 
The selected action $sa$, $sa\in A$;
\Function{MonteCarloSearch}{$time\_limit$}
\State sa\ =\ A.get(0);//default value of sa is set as the first action in $A$
\If{A.size()\ $>$\ 1}
\For{int i\ =\ 0; i\ $<$\ A.size(); i\ =\ (i+1)\%A.size()}
\If{time\_cost\ $>$\ time\_limit}
\State break;
\EndIf
\State a\ =\ A.get(i);
\State score\ =\ getGoalByPerformingRandomActionsFromNextState(s,a);
\State score[i]\ +=\ score;
\State visit[i]\ +=\ 1;
\EndFor
\State highest\_score\ =\ 0;
\State best\_action\_index\ =\ 0;
\For{int i\ =\ 0; i\ $<$\ A.size(); i++}
\State expected\_score[i]\ =\ score[i]/visit[i];
\If{expected\_score[i]\ $>$\ highest\_score}
\State highest\_score\ =\ expected\_score[i];
\State best\_action\_index\ =\ i;
\EndIf
\EndFor
\State sa\ =\ A.get(best\_action\_index)
\EndIf
\State\Return{sa}; 
\EndFunction
\end{algorithmic}
\end{algorithm}
\subsection{Inserting MCS inside Q-learning}
We will now add Monte Carlo Search to Q-learning (Algorithm 4). Starting from plain Q-learning,
in Algorithm~\ref{alg:algorithm4} (line~\ref{alg:algorithm4:random}), we see that a {\em random action\/} is chosen when QPlayer can not find an existing value in the $Q(s,a)$ table. In this case, QPlayer acts like a random game player, which will lead to a low win rate and slow learning speed. In order to address this problem, we introduce a variant of Q-learning combined with MCS. MCS performs a time limited lookahead for good moves. The more time it has, the better the action it finds will be. See Algorithm~\ref{alg:algorithm6} (line ~\ref{alg:algorithm6:mcs}).

By adding  MCS, we effectively add a local version of the last two stages of MCTS to Q-learning: the playout and backup stage~\cite{Browne2012}.

\allowdisplaybreaks
\begin{algorithm}[H]
\renewcommand{\algorithmicrequire}{ \textbf{Input:}} 
\renewcommand{\algorithmicensure}{ \textbf{Output:}} 
\caption{Monte Carlo Q-learning Player For Two-Player Zero-Sum Games}
\label{alg:algorithm6}
\begin{algorithmic}[1] 
\Require ~~\\ 
game state: \emph{S};\\
legal actions:\emph{A};\\
learning rate: $\alpha$;\\
discount factor: $\gamma$;\\
corresponding updating tables: $Q_{myrole}(S,A)$ for every role in the game;
\Ensure ~~\\ 
selected action according to updating table: $Q_{myrole}(S,A)$;
\Function{epsilonGreedyMonteCarloQlearning}{$S,A$}
\If{$\epsilon$-greedy is enabled}
\For {each learning match}
\State record\ =\ getMatchRecord();
\For{each state from termination to the beginning in record}
\State myrole\ =\ getCurrentRole();
\State R(s,a)\ =\ getReward(s,a);//$s^\prime$ is terminal state?\ getGoal($s^\prime$,myrole):0
\State Update $Q_{myrole}(s,a)\leftarrow (1-\alpha)Q_{myrole}(s,a)+\alpha(R(s,a)+\gamma max_{a^\prime}Q_{myrole}(s^\prime , a^\prime))$;
\EndFor
\EndFor
\State selected\ =\ false;
\State expected\_score\ =\ 0;
\For{each $q_{myrole}(s,a)$ in $Q_{myrole}(S,A)$}
\State if(current game state equals s and expected\_score\ $<\ q_{myrole}(s,a))$;
\State expected\_score\ =\ $q_{myrole}(s,a)$;
\State selected\_action\ =\ a;
\State selected\ =\ true;
\EndFor
\If{selected\ ==\ false}
\State \emph{\textbf{selected\_action\ =\ MonteCarloSearch(time\_limit)}};\label{alg:algorithm6:mcs} // Algorithm~\ref{alg:algorithm1}
\EndIf
\Else
\State selected\_action\ =\ Random()
\EndIf
\State \Return selected\_action; 
\EndFunction
\end{algorithmic}
\end{algorithm}

\subsubsection*{Experiment 2} 
\begin{figure}[H]
\centering
\subfigure[total learning match=5000]{\label{fig:subfig4:a} 
\includegraphics[width=0.48\textwidth]{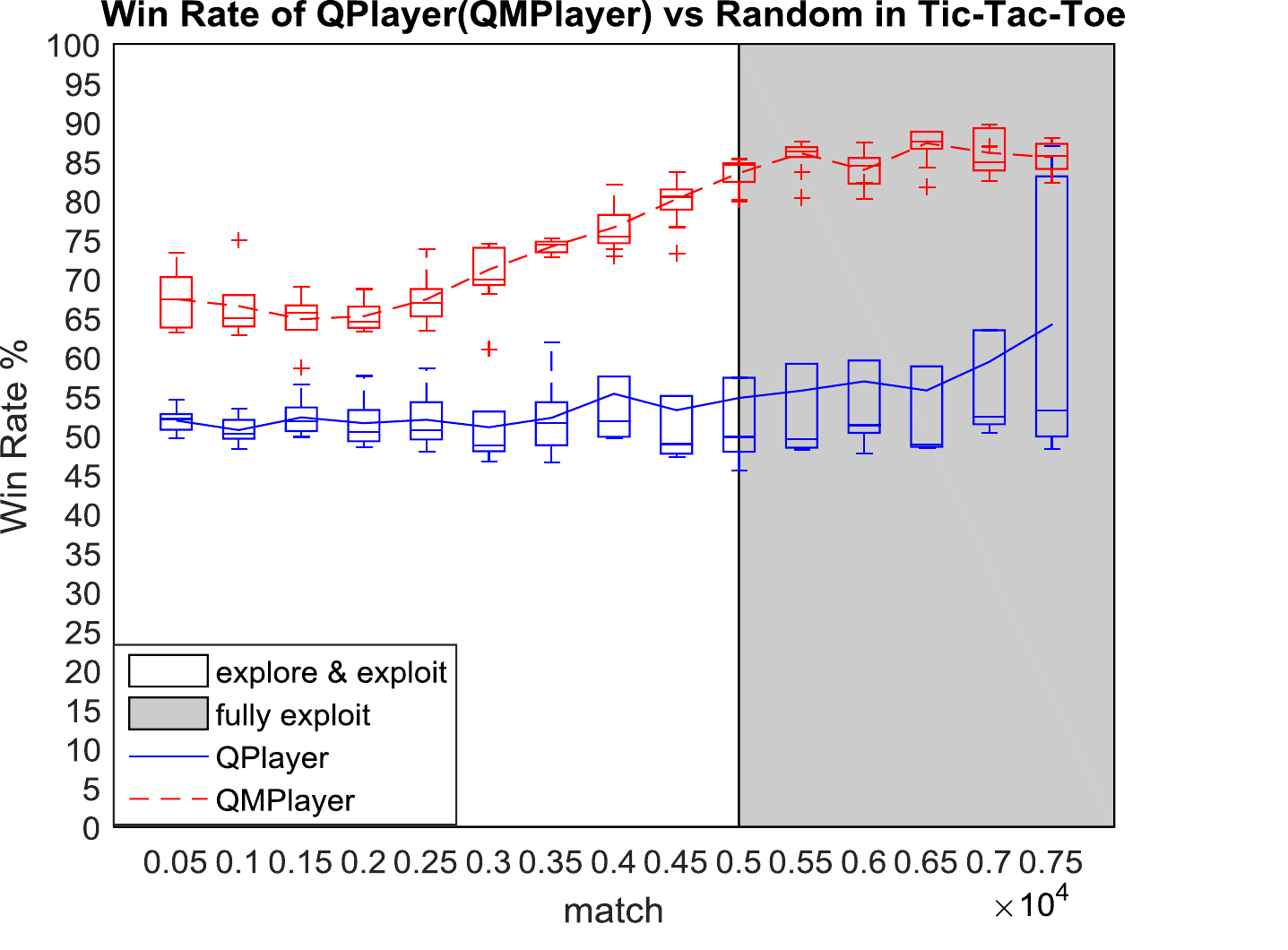}}
\hspace{0.000000000000001\textwidth}
\subfigure[total learning match=10000]{\label{fig:subfig4:b} 
\includegraphics[width=0.48\textwidth]{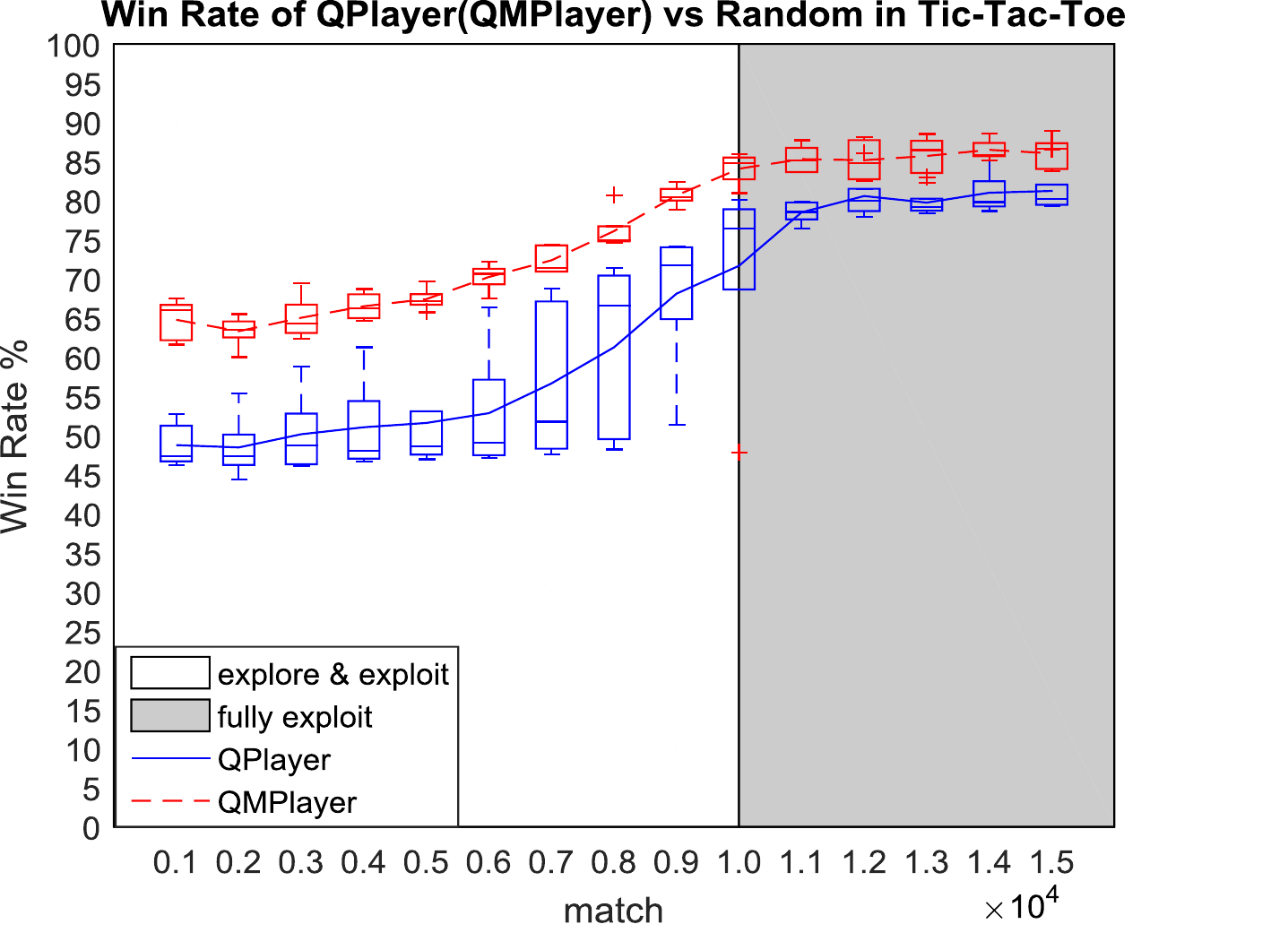}}
\hspace{0.000000000000001\textwidth}
\subfigure[total learning match=20000]{\label{fig:subfig4:c} 
\includegraphics[width=0.48\textwidth]{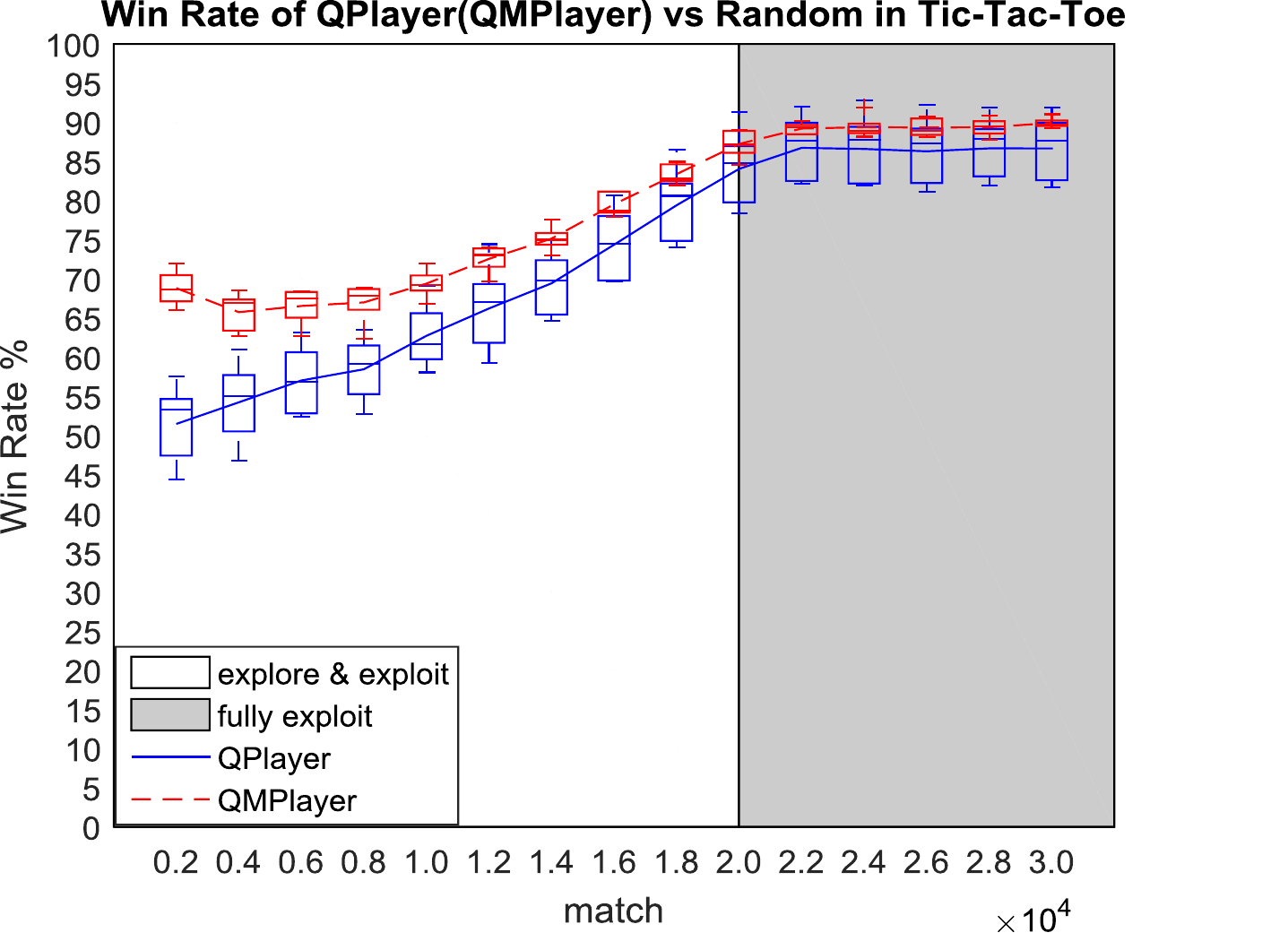}}
\hspace{0.000000000000001\textwidth}
\subfigure[total learning match=30000]{\label{fig:subfig4:d} 
\includegraphics[width=0.48\textwidth]{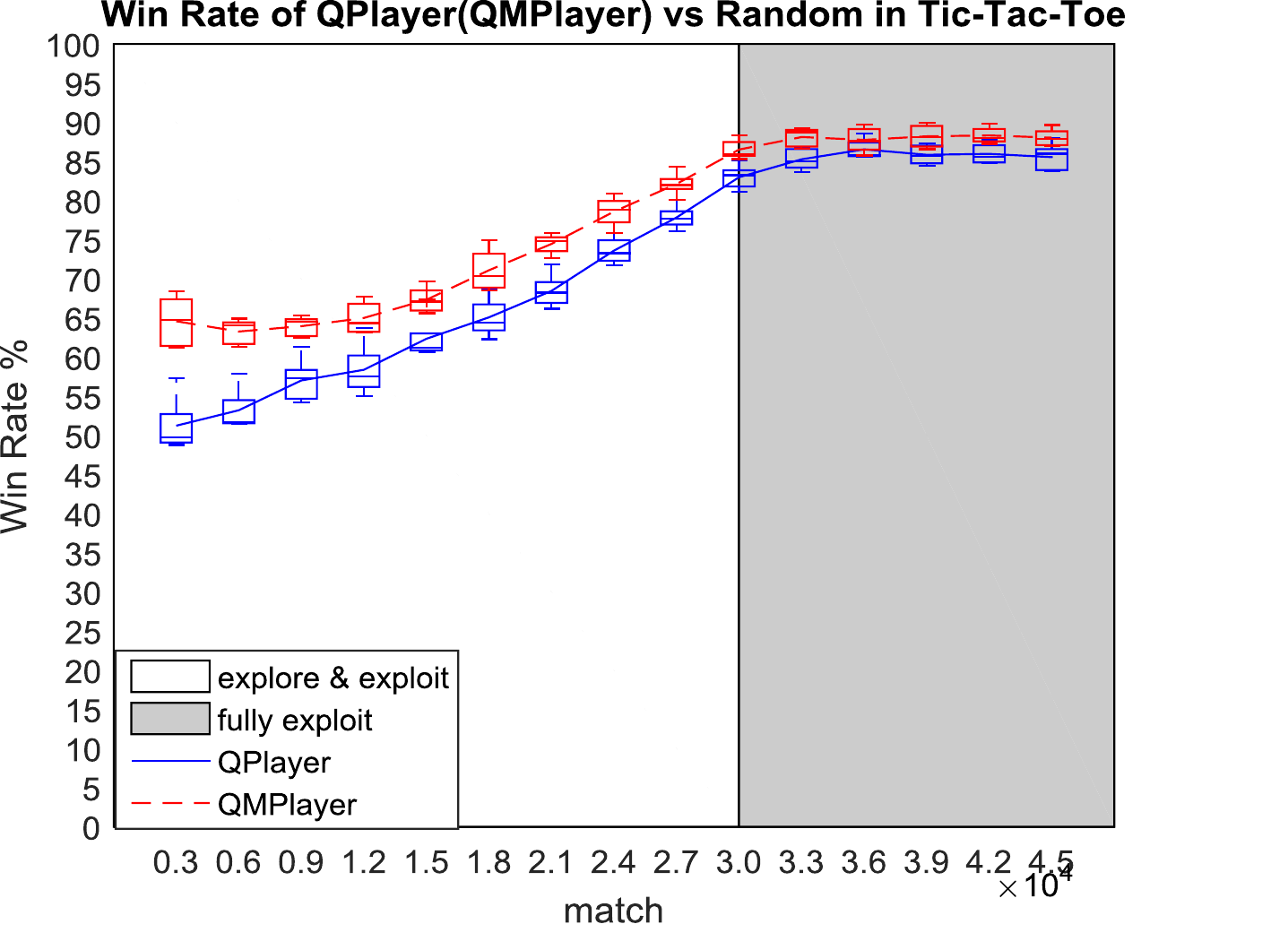}}
\hspace{0.000000000000001\textwidth}
\subfigure[total learning match=40000]{\label{fig:subfig4:e} 
\includegraphics[width=0.48\textwidth]{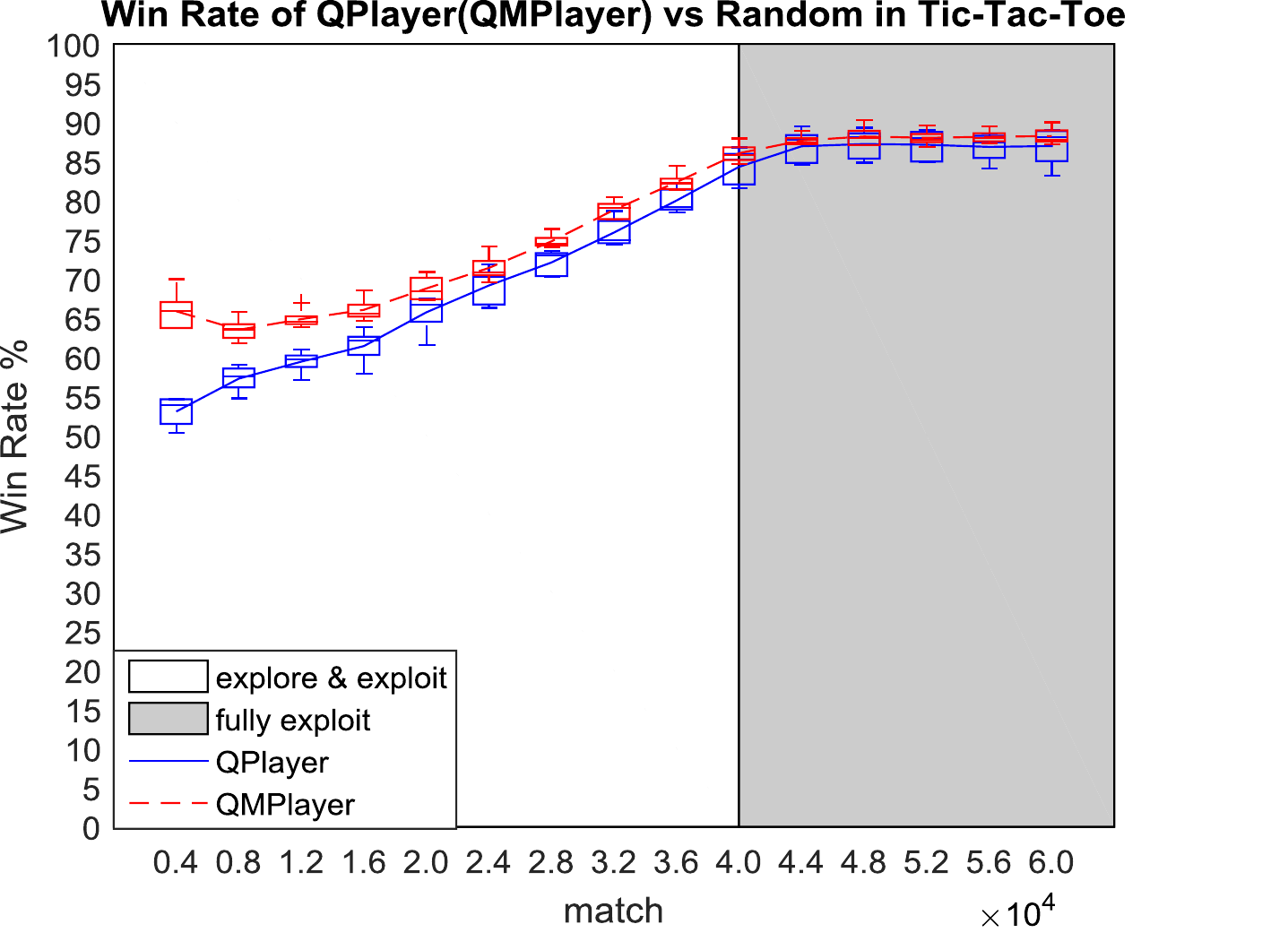}}
\hspace{0.000000000000001\textwidth}
\subfigure[total learning match=50000]{\label{fig:subfig4:f} 
\includegraphics[width=0.48\textwidth]{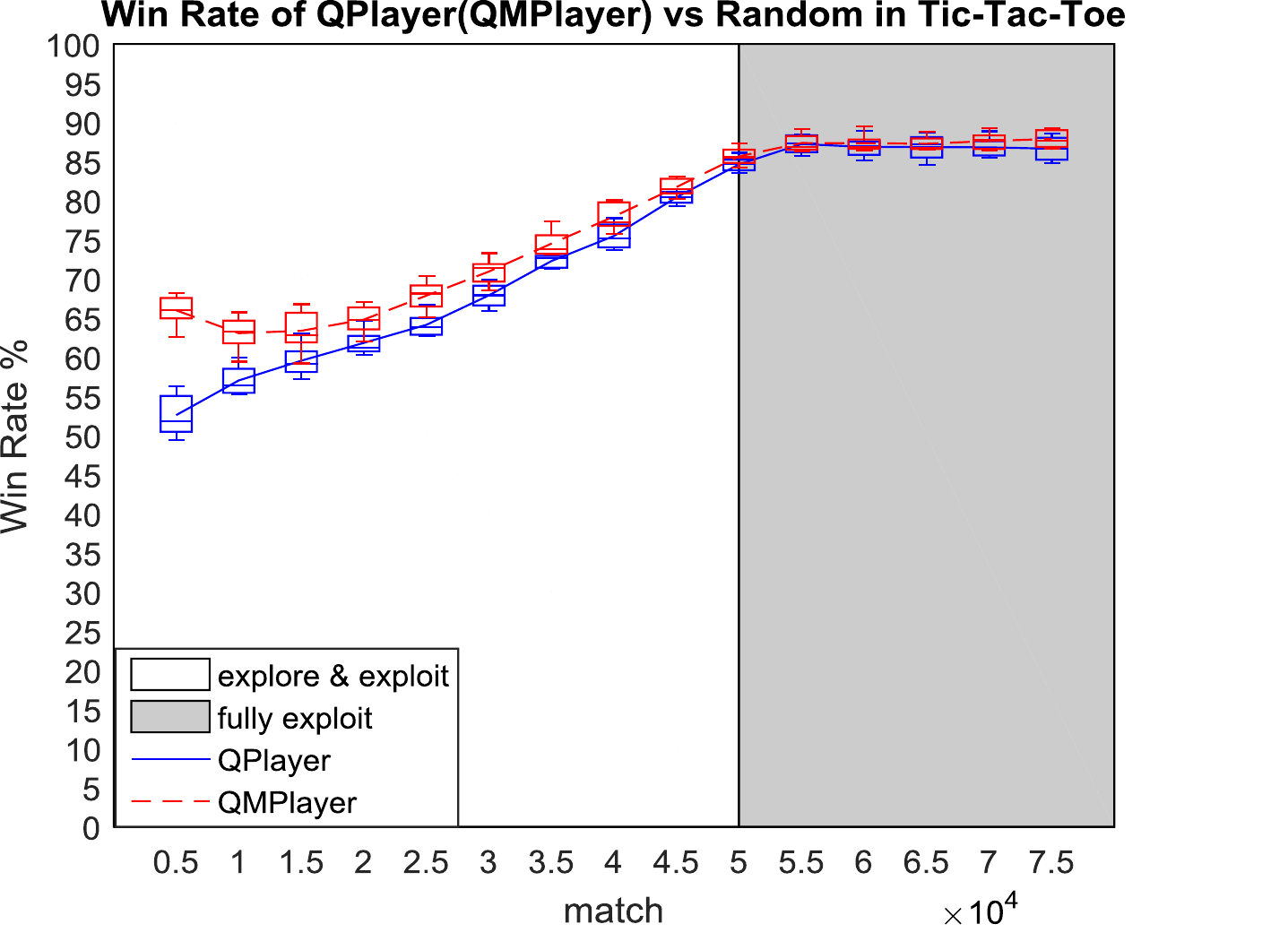}}
\caption{Win Rate of QMPlayer and QPlayer vs Random in Tic-Tac-Toe for 5 experiments. Small Monte Carlo lookaheads improve the convergence of Q-learning, especially in the early part of the learning. QMPlayer always outperforms Qplayer}
\label{fig:subfig4} 
\end{figure}

We will now describe our second experiment. In this experiment, with Monte Carlo-enhanced Q-learning, we use the QMPlayer. See Algorithm~\ref{alg:algorithm6}. We set parameters $\alpha=0.1$, $\gamma=0.9$, $\epsilon \in[0, 0.5]$, $time\_limit=50 ms$ respectively.
For QMPlayer to learn to play Tic-Tac-Toe, we also set the {\em total learning match}=5000, 10000, 20000, 30000, 40000, 50000,  respectively, and then make it play the game with Random player for $1.5\ \times\ ${\em total learning match} matches for 5 rounds. The comparison with QPlayer is shown in Fig.\ref{fig:subfig4}.
In Fig.\ref{fig:subfig4:a}, QMPlayer gets a high win rate(about 67\%) at the very beginning. Then the win rate decreases to 66\% and 65\%, and then increases from 65\% to around 84\% at the end of $\epsilon$ learning(match=5000). Finally, the win rate stays at around 85\%. Also in the other sub figures, for QMPlayer, the trend of all curves  decreases first and then increase until reaching a stable state. This is because at the very beginning, QMPlayer chooses more actions from MCS. Then as the learning period moves forward, it chooses more actions from Q table and achieves convergence.

\begin{figure}[H]
\centering
\includegraphics[width=0.65\textwidth]{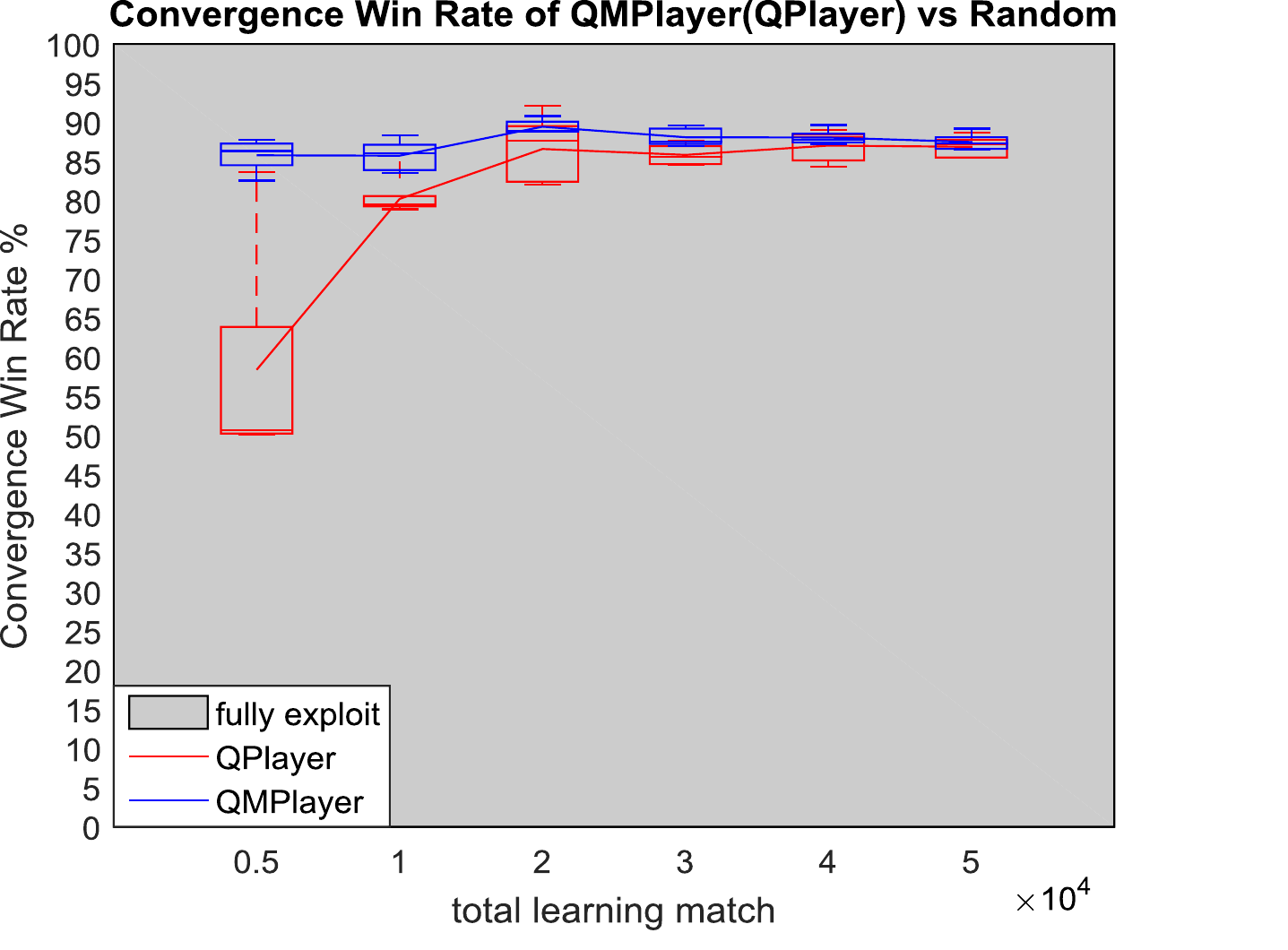}
\caption{Convergence Win Rate of QMPlayer and QPlayer vs Random in Tic-Tac-Toe}
\label{fig:fig3} 
\end{figure}

Note that in every sub figure, QMPlayer can always achieve a higher win rate than QPlayer,
not only at the beginning but also at the end of the learning period. Overall, QMPlayer achieves a better performance than QPlayer with the higher convergence win rate (at least 87.5\% after training 50000 matches). To compare the convergence speeds of QPlayer and QMPlayer, we summarize the convergence win rates of different \emph{total learning match} according to Fig.~\ref{fig:subfig2} and Fig.~\ref{fig:subfig4}, in Fig.\ref{fig:fig3}.

These results show that combining MCS with Q-learning for GGP can improve the win rate both at the beginning and at the end of the learning period. The main reason is that Monte Carlo-enhanced Q-learning allows the $Q(s,a)$ table to be filled quickly with good actions from MCS, achieving a quick and direct learning rate. It is worth to note that,  QMPlayer will spend slightly more time (at most is $search~time~limit\times$ {\em number~of~(state~action)~pairs}) in training than QPlayer.
\subsection{Comparison with MCTS}
In order to evaluate the performance of both Q-learning players, we  implemented a basic MCTS player \cite{Mehat2008}. Referring to the basic MCTS algorithm in~\cite{Browne2012}, we present the pseudo code of basic time limited MCTS in GGP in Algorithm 5.

\allowdisplaybreaks
\begin{algorithm}[H]
\renewcommand{\algorithmicrequire}{ \textbf{Input:}} 
\renewcommand{\algorithmicensure}{ \textbf{Output:}} 
\caption{Basic Time Limited Monte Carlo Tree Search Player Algorithm For Two-Player Zero-Sum Games}
\label{alg:algorithm5}
\begin{algorithmic}[1] 
\Require ~~\\ 
game state: \emph{S};\\
legal actions:\emph{A};\\
empty game tree:\emph{tree};\\
visited nodes: \emph{visited};\\
current node:\emph{node};
\Ensure ~~\\ 
selected action according to updated game tree;
\Function{MCTS}{$S,A,time\_limit$}
\If{legal\_moves.size()\ ==\ 1}
\State selected\_ action\ =\ legal\_moves.get(0);
\Else
\While{time\_cost $\leq$ time\_limit}
\While{!node.isLeaf()}
\State node\ =\ selectNodeByUCTMinMax();
\State visited.add(node);
\EndWhile
\State expandGameTree(); //expand tree based on the number of all next states
\State node=selectNodeByUCTMinMax();
\State visited.add(node);
\State bonus\ =\ playout(); //simulate from node to terminal state, get a score.
\State backUpdate(); //for every visited node, count+=1; value+=bonus;
\State visited.removeAll(visited);//erase visited list
\EndWhile
\State selected\_child=getChildWithMaxAverageValue(tree.get(0).children)
\State selected\_action=getMoveFromParentToChild(selected\_child);
\EndIf
\State \Return selected\_action; 
\EndFunction
\Function{selectNodeByUCTMinMax}{}
\For{each child in node.children}
\State float uct\ =\ $\frac{child.totalvalue}{child.visitcount}+\sqrt{\frac{ln(node.visitcount+1)}{child.visitcount}}$;
\If{is my turn according to node.game\_state}
\If{max\_value\ $<$\ uct}
\State max\_value\ =\ uct;
\State selected\_node\ =\ child;
\EndIf
\Else
\If{min\_value\ $>$\ uct}
\State min\_value\ =\ uct;
\State selected\_node\ =\ child;
\EndIf
\EndIf
\EndFor
\State \Return selected\_node;
\EndFunction
\end{algorithmic}
\end{algorithm}

First, we make QPlayer learn 50000 matches. Then we set \emph{time\_limit}=10s for the MCTS player to build and update the search tree. For MCS, we also allow 10 seconds. With this long time limit, they reach perfect play on this small game. QPlayer and QM-player, in contrast, only get 50ms MCS time, and cannot reach perfect play in this short time period. QPlayer plays against the MCTS player in GGP by playing Tic-Tac-Toe for 100 matches. Then we pit other players against each other. The most relevant competition results of different players mentioned in this paper are shown in Table~\ref{tab1}. The cells contain the win rate of the column player against the row player.
\begin{table}
\centering
\begin{tabular}{|l|l|l|l|l|l|}
\hline
     &  MCTS & Random & QPlayer & QMPlayer & MCS\\
\hline
  MCTS & - & 0.5\% & 0 & 0 & 35\%\\
\hline
 Random & 99.5\% & - & 86.5\% & 87.5\% & 100\%\\
\hline
  QPlayer & 100\% & 13.5\% & - & - & - \\
\hline
  QMPlayer &  100\% & 12.5\% & - & - & -\\
  \hline
  MCS &  65\% & 0 & - & - & - \\
\hline
\end{tabular}
\linebreak
\caption{Summary of Win Rate of Different Players Against to Each Other. The state space of Tic-Tac-Toe is too small for MCTS, it reaches perfect play. QMPlayer out-performs QPlayer}\label{tab1}
\end{table}

In Table~\ref{tab1}, we find that (1) the state space of Tic-Tac-Toe is too small for MCTS, which reaches perfect play (win rate of 100\%). Tic-Tac-Toe is suitable for showing the difference between QPlayer and QMPlayer. (2) MCTS wins 65\% matches against QMPlayer since MCTS can win in the first hand matches and always get a draw in the second hand matches while playing with MCS. (3) The convergence win rate of QMPlayer(87.5\%) against to Random is slightly higher than QPlayer(86.5\%).

\section{Conclusion}
This paper examines the applicability of Q-learning, the canonical reinforcement learning method, to create general algorithms for GGP programs. Firstly, we show how good canonical implementations of Q-learning perform on GGP games. The GGP system allows us to easily use three real games for our experiments: Tic-Tac-Toe, Connect Four, and Hex. We find that (1) Q-learning is indeed general enough to achieve convergence in GGP games.
With Banerjee~\cite{Banerjee2007}, however, we also find that (2) convergence is slow. Compared against the MCTS algorithm that is often used in GGP~\cite{Mehat2008}, performance of Q-learning is lacking: MCTS achieves perfect play in Tic-Tac-Toe, whereas Q-learning does not.

We then enhance Q-learning with an MCS based lookahead. We find that, especially at the start of the learning, this speeds up convergence considerably. Our Q-learning is table-based, limiting it to small games. Even with the MCS enhancement, convergence of QM-learning does not yet allow its direct use in larger games.
The QPlayer needs to learn a large number of matches to get good performance in playing larger games. The results with the improved Monte Carlo algorithm (QM-learning) show a real improvement of the player's win rate, and learn the most probable strategies to get high rewards faster than learning completely from scratch.

A final result is, that, where Banerjee et al. used a static value for $\epsilon$, we find that a value for $\epsilon$ that changes with the learning phases gives better performance (start with more exploration, become more greedy later on).

The table-based implementation of Q-learning facilitates theoretical analysis, and comparison against some baselines \cite{Banerjee2007}. However, it is only suitable for small games. A neural network implementation facilitates the study of larger games, and allows meaningful comparison to DQN variants \cite{Mnih2015}.

Our  use of Monte Carlo in QM-learning is different from the AlphaGo architecture, where MCTS is wrapped around Q-learning (DQN) \cite{Mnih2015}. In our approach, we inserted Monte Carlo {\em within} the Q-learning loop.
Future work should show if our QM-learning results transfer to AlphaGo-like uses of DQN inside MCTS, if QM-learning can  achieve faster convergence, reducing the high computational demands of AlphaGo \cite{Silver2017a}. Additionally, we plan to study  nested MCS in Q-learning \cite{Cazenave2016}. Implementing Neural Network based players also allows the study of more complex GGP games.

\subsubsection*{Acknowledgments.} Hui Wang acknowledges financial support from the China Scholarship Council (CSC), CSC No.201706990015.

\end{document}